\documentclass[sigconf]{acmart}
\AtBeginDocument{%
  }

\copyrightyear{2025}
\acmYear{2025}
\setcopyright{acmlicensed}\acmConference[MM '25]{Proceedings of the 33rd ACM International Conference on Multimedia}{October 27--31, 2025}{Dublin, Ireland}
\acmBooktitle{Proceedings of the 33rd ACM International Conference on Multimedia (MM '25), October 27--31, 2025, Dublin, Ireland}
\acmDOI{10.1145/3746027.3755524}
\acmISBN{979-8-4007-2035-2/2025/10}
\begin{document}

\title{AnimeColor: Reference-based Animation Colorization with Diffusion Transformers}

\settopmatter{authorsperrow=4}
\author{Yuhong Zhang}
\affiliation{%
  \institution{Shanghai Jiao Tong University}
  \city{Shanghai}
  \country{China}}
\email{rainbowow@sjtu.edu.cn}

\author{Liyao Wang}
\affiliation{%
  \institution{Tianjin University}
  \city{Tianjin}
  \country{China}}
\email{wly_tju@126.com}

\author{Han Wang}
\affiliation{%
  \institution{Shanghai Jiao Tong University}
  \city{Shanghai}
  \country{China}}
\email{esmuellert@sjtu.edu.cn}

\author{Danni Wu}
\affiliation{%
  \institution{Communication University of China}
  \city{Beijing}
  \country{China}}
\email{danniwu@cuc.edu.cn}

\author{Zuzeng Lin}
\affiliation{%
  \institution{Tianjin University}
  \city{Tianjin}
  \country{China}}
\email{linzuzeng@tju.edu.cn}

\author{Feng Wang}
\affiliation{%
  \institution{CreateAI}
  \city{Beijing}
  \country{China}}
\email{feng.wff@gmail.com}

\author{Li Song}
\authornote{Corresponding author.}
\affiliation{%
  \institution{Shanghai Jiao Tong University}
  \city{Shanghai}
  \country{China}}
\email{song\_li@sjtu.edu.cn}

\renewcommand{\shortauthors}{Yuhong Zhang et al.}

\begin{abstract}
Animation colorization plays a vital role in animation production, yet existing methods struggle to achieve color accuracy and temporal consistency. To address these challenges, we propose \textbf{AnimeColor}, a novel reference-based animation colorization framework leveraging Diffusion Transformers (DiT). Our approach integrates sketch sequences into a DiT-based video diffusion model, enabling sketch-controlled animation generation. We introduce two key components: a High-level Color Extractor (HCE) to capture semantic color information and a Low-level Color Guider (LCG) to extract fine-grained color details from reference images. These components work synergistically to guide the video diffusion process. Additionally, we employ a multi-stage training strategy to maximize the utilization of reference image color information. Extensive experiments demonstrate that AnimeColor outperforms existing methods in color accuracy, sketch alignment, temporal consistency, and visual quality. Our framework not only advances the state of the art in animation colorization but also provides a practical solution for industrial applications. The code will be made publicly available at \href{https://github.com/IamCreateAI/AnimeColor}{https://github.com/IamCreateAI/AnimeColor}.

\end{abstract}


\begin{teaserfigure}
  \centering
  \vspace{-3mm}
  \includegraphics[width=0.80\textwidth]{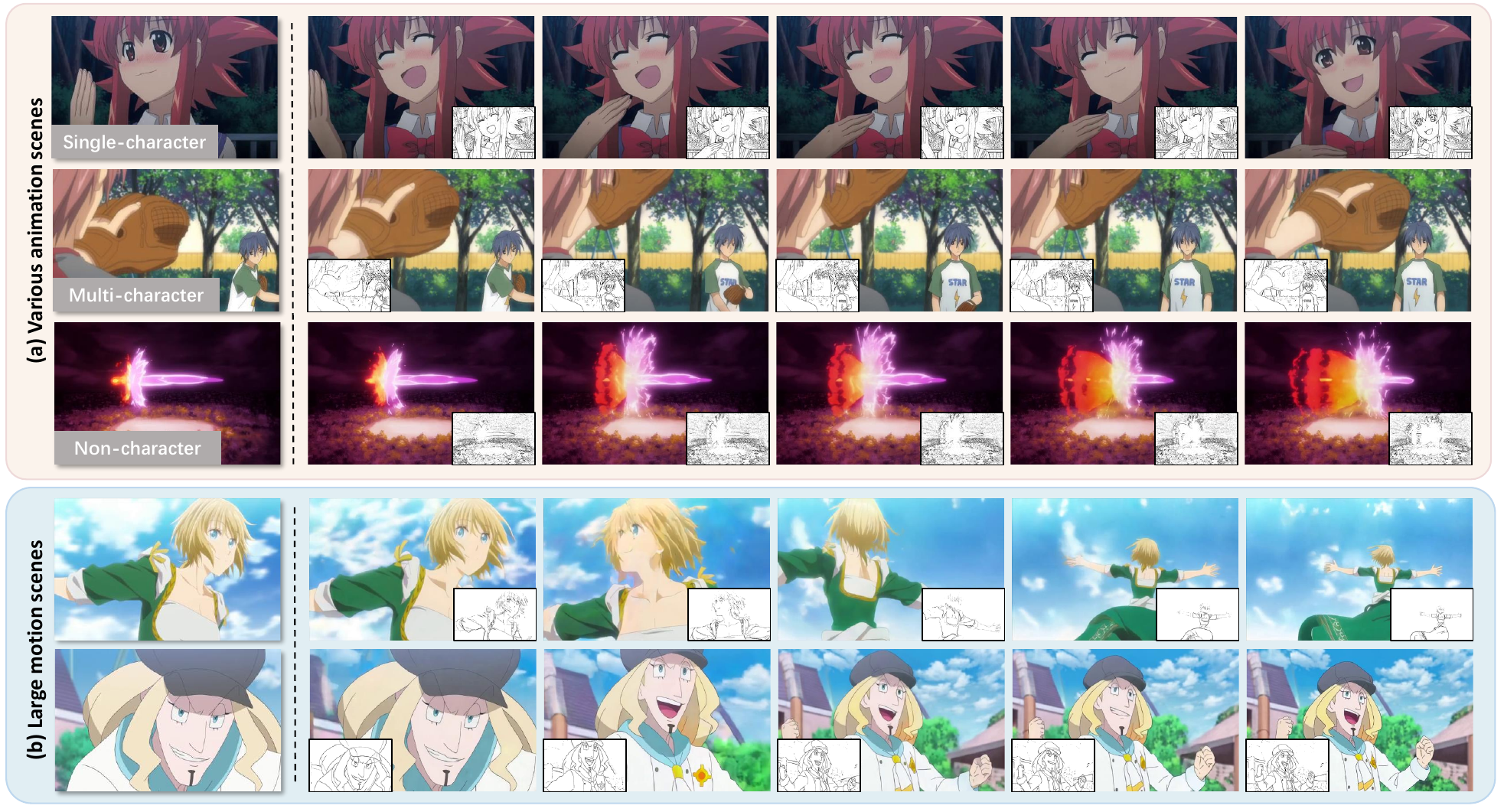}
  \vspace{-3mm}
  \caption{Given a reference image and a sequence of sketches, our method generates high-quality colorized animation videos. AnimeColor demonstrates robust performance across diverse animation scenes, including single-character, multi-character, and non-character scenarios. Notably, it excels in challenging large-motion scenes, such as back or side views and significant character movements, where existing methods often struggle.}
  \label{fig:teaser}
\end{teaserfigure}


\maketitle

\section{Introduction}
Animation has become a ubiquitous medium for storytelling and artistic expression, yet its production remains highly labor-intensive. A critical step in this process is colorization, where animators painstakingly colorize keyframes and extend these colors to all frames while maintaining style and temporal consistency. Automating this task is essential for reducing costs and accelerating content creation.

Previous works~\cite{shi2022referencevc,dou2021dualcolor,zhang2021cmft} have attempted to automate colorization. However, they often handle each frame separately, leading to flickering and inconsistency that can affect the visual appeal of the animation. Even when incorporating previous frames as references for processing subsequent frames~\cite{thasarathan2019tcvc}, error accumulation can occur, diminishing the viewer's experience. Recently, video generation has achieved great success due to the advancement of diffusion models~\cite{chen2023videocrafter1,blattmann2023svd, yang2024cogvideox}. Researchers are exploring integrating diffusion models into the animation colorization task, utilizing the prior knowledge from visual generative models to capture coherent visual features and ensure temporal consistency. LVCD~\cite{huang2024lvcd} has proposed a method for handling long video animation colorization, and AniDoc~\cite{meng2024anidoc} has introduced a technique for character animation colorization. However, these methods have difficulty in addressing cases involving large motion and non-character scenes. Tooncrafter~\cite{xing2024tooncrafter} also achieves controllable video generation with sketches and keyframes but exhibits noticeable flickering and poor color consistency. Meanwhile, DiT-based architectures~\cite{yang2024cogvideox,li2024hunyuan} have demonstrated superior visual quality and temporal consistency in video generation, yet their potential for animation colorization remains largely unexplored. This challenge may stem from the inherent limitations of DiT compared to U-Net, including greater training convergence difficulties and control complexities.

To address these limitations, we propose \textbf{AnimeColor}, a novel reference-based animation colorization framework built on Diffusion Transformers (DiT). Our key innovation lies in the integration of sketch sequences into a DiT-based video diffusion model, enabling sketch-controlled animation generation. To ensure precise color control, we introduce two complementary modules: a High-level Color Extractor (HCE) for semantic color consistency and a Low-level Color Guider (LCG) for fine-grained color accuracy. These modules jointly guide the diffusion process, addressing both semantic alignment and detail-level precision. Additionally, we design a multi-stage training strategy to maximize the utilization of reference image information while maintaining temporal consistency. 

\begin{figure}[]
  \centering
    \includegraphics[width=0.9\linewidth]{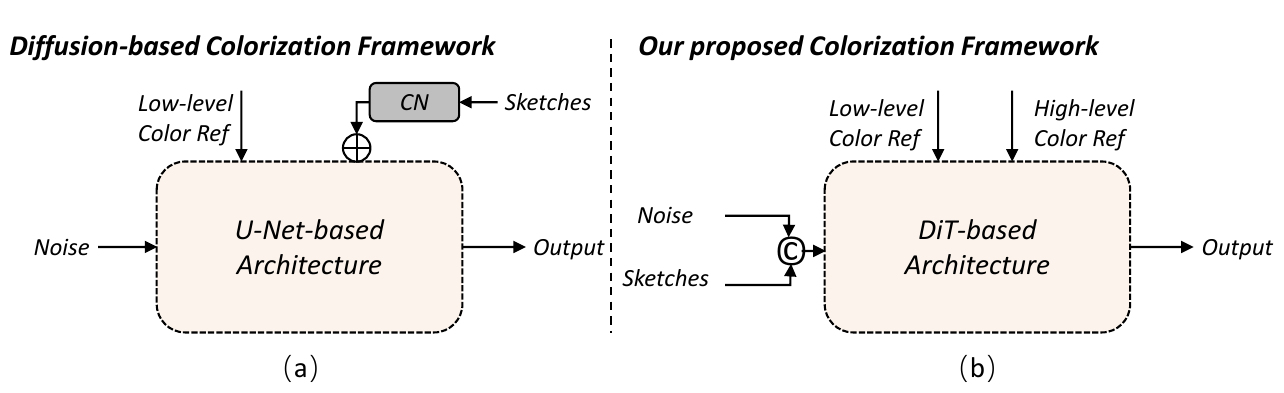}
  \vspace{-4mm}
  \caption{Illustration of the difference between our proposed AnimeColor and other diffusion-based animation colorization methods (CN denotes ControlNet). In our method, high-level color reference and low-level color reference complement each other to improve the accuracy of the generated animation. Sketch control is introduced by concatenating to better adapt the video generation model of the DiT-based architecture, reduce the difficulty of training, and improve sketch alignment. What’s more, the DiT-based architecture with greater generation potential contributes to the temporal consistency.}
  \label{fig:introduction_compare}
\end{figure}

AnimeColor distinguishes itself from existing methods in three key aspects (see Fig. \ref{fig:introduction_compare}): (1) the combination of HCE and LCG for robust color control, (2) sketch conditioning through latent concatenation for improved alignment, and (3) a more robust DiT-based video generation model for enhanced temporal consistency. As shown in Fig. \ref{fig:teaser}, our method excels in diverse scenarios, including single-character, multiple-character and non-character scenes. What's more, our method shows excellent performance in the challenging large-motion scenes, including both camera movement and character movement. Extensive qualitative and quantitative experiments validate AnimeColor’s superiority in color accuracy, sketch alignment, and temporal consistency.

Our contributions are summarized as follows:
\begin{itemize}
    \item We introduce AnimeColor, a DiT-based framework for animation colorization that significantly improves color accuracy and temporal consistency, particularly in large-motion scenes.
    \item We propose a complement color control mechanism, combining HCE and LCG, to achieve both semantic and fine-grained color accuracy.
    \item Comprehensive experiments demonstrate the superiority of AnimeColor in color accuracy, sketch alignment, temporal consistency, and visual quality.
\end{itemize}

\section{Related Work}
\subsection{Video Generation Models.}
With the advancement of image generation techniques based on diffusion models~\cite{dhariwal2021diffusionbeatgan,rombach2022ldm}, a plethora of video generation models have emerged. From a model architecture perspective, there are primarily two types of architectures for video generation models. Primitive video diffusion models, such as VideoCrafter~\cite{chen2023videocrafter1}, SVD~\cite{blattmann2023svd}, and some other models~\cite{guo2023animatediff, xing2024tooncrafter, yang2025layeranimate}, achieve this by incorporating temporal self-attention into the U-Net diffusion backbone. Recently, with the increase in model complexity and data scale, video generation models based on Diffusion Transformers (DiT) have been proposed~\cite{xu2024easyanimate, li2024hunyuan, yang2024cogvideox}. For instance, CogVideoX~\cite{yang2024cogvideox} introduces an expert DiT with stacked 3D attention blocks dedicated to concatenating context embeddings and visual tokens. Video generation models based on DiT demonstrate significant generative potential~\cite{zheng2024opensora}, yet research on their controllability remains limited. As generated videos become increasingly realistic, imparting control over the generation process becomes crucial. Leveraging the DiT-based video architecture, we introduce the reference-based animation colorization framework.

\subsection{Lineart Image Colorization.}
The objective of lineart image colorization is to accurately fill the blank spaces within the lineart images with appropriate colors. Unlike grayscale images, line drawings consist solely of structural outlines, devoid of details such as brightness or texture. This absence of detail introduces significant challenges to the lineart image colorization process. Traditional methods of coloring line drawings have been semi-automatic, involving user interaction to apply colors to designated areas~\cite{chen2020active}. However, the advent of deep learning has catalyzed significant advancements in this domain. Modern techniques now facilitate lineart image colorization with the aid of color hint points~\cite{zhang2018two}, text labels~\cite{kim2019tag2pix}, or natural language inputs~\cite{zou2019language}. Another notable approach is reference-based lineart image colorization, where users provide a reference image to guide the colorization of the line drawing in a similar style and palette. BasicPBC~\cite{dai2024learning} has introduced a novel learning-based approach that incorporates a matching process and has also contributed a PaintBucket-Character dataset. More recently, the substantial enhancements in generative model capabilities have prompted extensive exploration of diffusion models for lineart image colorization. ColorizeDiffusion~\cite{yan2024colorizediffusion} offers solutions to potential conflicts between image hints and sketch inputs, while AnimeDiffusion~\cite{cao2024animediffusion} presents an automated method for lineart image colorization of anime facial line drawings through a hybrid training strategy. 

\subsection{Reference-based sketch-guided Video Colorization.}
Reference-based colorization is the most common and important method in animation colorization. CMFT~\cite{zhang2021cmft} uses animation frame pairs for training to achieve the sketch's colorization, but this method only ensures consistency with the reference frame and does not consider temporal consistency. TCVC~\cite{thasarathan2019tcvc} uses the previous colorized frame as a reference to guide subsequent frame colorization to ensure short-term consistency, but it may lead to error accumulation. TRE-Net~\cite{wang2023trenet} uses the first reference frame and the previous frame to solve the error accumulation problem. ACOF~\cite{yu2024acof} proposes a color propagation mechanism based on optical flow, but it still requires refinement and there are also error accumulation problems. With the advancement of the generative model, LVCD~\cite{huang2024lvcd} proposes an animation colorization framework based on video diffusion, but this method is based on SVD~\cite{blattmann2023svd}, lacks text control and semantic-level color control, and does not work well in some large-scale motion scenes. AniDoc~\cite{meng2024anidoc} proposes an animation colorization method based on character reference, but this method is more suitable for character colorization and doesn't perform well on background colorization. ToonCrafter~\cite{xing2024tooncrafter} is an animation interpolation model that can also be used for colorization based on reference frames and sketches, but it is short in temporal consistency as well as color consistency with reference images. In addition, these diffusion-based methods all build on the video generation model of the U-Net architecture. We build on the more generative video generation model of the DiT-based architecture and introduce more precise color control, improving both color consistency with the reference image and temporal consistency.

\section{Method}

\begin{figure*}[h]
  \centering
  \includegraphics[width=0.80\linewidth]{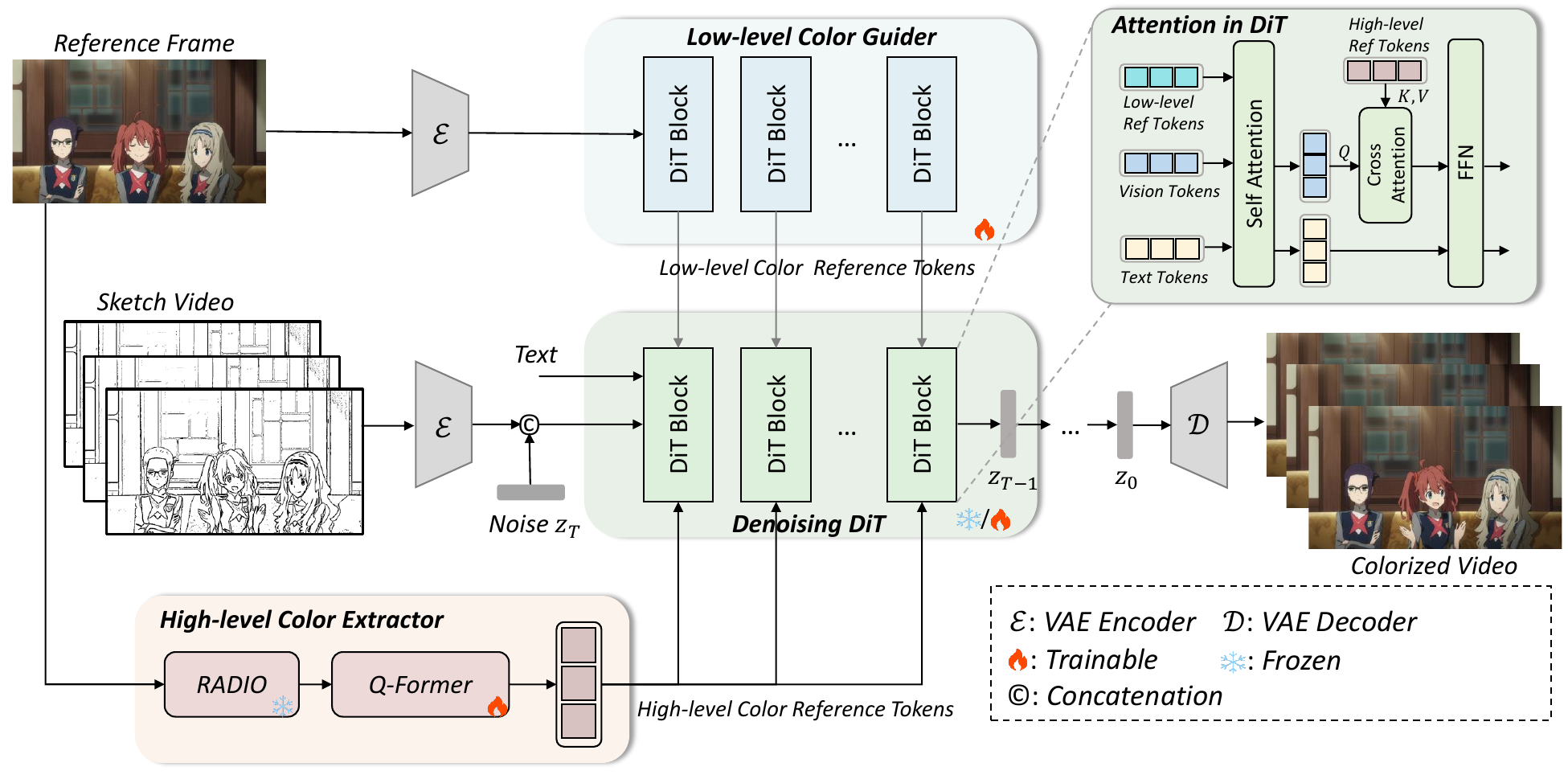}
  \vspace{-3mm}
  \caption{The framework of our proposed AnimeColor. We first concatenate sketch latents with noise as the input for the denoising DiT. Subsequently, we introduce the High-level Color Extractor and Low-level Color Guider to extract color information from the reference image for controlling video generation. The color control is integrated into the denoising DiT through attention blocks. Additionally, we design a four-stage training strategy to ensure the effectiveness of each module.}
  \label{fig:framework}
\end{figure*}

\subsection{Overview}
The framework of our AnimeColor is illustrated in Fig. \ref{fig:framework}. Given a reference image $I_{\rm ref}$ and a sequence of sketches $S=\{I_{\rm skt}^{1},...,I_{\rm skt}^{N}\}$, our goal is to generate a temporally consistent video sequence $G=\{I^{1},...,I^{N}\}$ that maintains structural consistency with $S$ while aligning in color and style with $I_{\rm ref}$. To achieve this, we first concatenate the sketch latents encoded by VAE Encoder with noised latent as the input of video diffusion transformers (DiTs). Thanks to the in-context learning capability of the DiT architecture, we can achieve sketch-controlled video generation. 

Secondly, in order to maintain color consistency between the generated video and the reference image, we design a \textbf{High-level Color Extractor (HCE)} and a \textbf{Low-level Color Guider (LCG)}. The High-level Color Extractor extracts high-level semantic color information, while the Low-level Color Guider extracts fine-grained color details. The two modules collectively control the color and style of the generated animation.

We integrate these color controls through the attention mechanism. The attention block in DiT is shown in the upper right corner of Fig. \ref{fig:framework}. We concatenate the low-level reference tokens with vision tokens and text tokens and perform self-attention to realize fine-grained color control. Then we perform the cross-attention between vision tokens and high-level reference tokens to realize semantic color control and avoid local color unevenness.

Finally, we implement a multi-stage training strategy to enable each module to fulfill its specific role effectively. The designed approach has empowered us to generate animated videos that stay aligned with the sketches, maintain color consistency with the reference images, and exhibit temporal stability.

\subsection{High-level Color Extractor}
With the inherent in-context learning~\cite{huang2024iclora} and generative ability of video diffusion transformers, our model can already achieve sketch-guided video generation. However, the color of the generated video is random and lacks accurate control. Therefore, we propose a \textbf{High-level Color Extractor (HCE)} to extract color control from $I_{\rm ref}$.

Previous studies~\cite{ye2023ipadapter} use the CLIP image encoder~\cite{radford2021clip} to extract image features to guide image generation. However, due to the limitations of CLIP's visual expression capabilities, the features it extracts are relatively limited, especially for animated videos. Therefore, we choose to use a more powerful visual foundation model RADIO~\cite{ranzinger2024radio} to encode the reference image. RADIO distills the knowledge of many foundation models such as CLIP~\cite{radford2021clip}, DINOv2~\cite{oquab2023dinov2}, and SAM~\cite{kirillov2023sam}, and has more powerful visual capabilities. It can help us extract more effective color information from the reference image. The process can be described as this:
\begin{equation}
    F_{\rm sum}, F_{\rm spa} = \mathcal{E}_{\rm RADIO} (I_{\rm ref}).
\end{equation}

By default, RADIO can return a summary feature $F_{\rm sum}$ and spatial features $F_{\rm spa}$. $F_{\rm sum}$ represents the general concept and $F_{\rm spa}$ represent the more localized feature. We aggregate high-level semantic information in $F_{\rm sum}$ and $F_{\rm spa}$ through MLP layers and generate the image embedding $F_{\rm RADIO}$.

In order to avoid the influence of content information in $F_{\rm RADIO}$, we use Q-Former~\cite{li2023blip} to aggregate image features to extract more effective information about color. We create $N$ learnable color tokens $F_{\rm query} \in \mathbb{R}^{N \times C}$, then concatenate them with $F_{\rm RADIO}$ for self-attention~\cite{vaswani2017attention}. We take the first $N$ tokens as high-level color reference  $F_{H} \in \mathbb{R}^{N \times C}$:
\begin{equation}
    F_{H} = \texttt{Self-Attention}(\texttt{Concat}(F_{\rm query}, F_{\rm RADIO})) [:N].
\end{equation}

This design ensures that the model effectively captures the color information in the reference images while filtering out irrelevant content details. After extracting intrinsic high-level color reference tokens, we employ cross-attention block to interact with the denoising DiT:
\begin{equation}
    F_{\rm out} = F_{\rm in} + \texttt{Cross-Attention}(F_{\rm in}, F_{H}),
\end{equation}
where $F_{\rm in}$ and $F_{\rm out}$ represent the input and output of the cross-attention module respectively.

\subsection{Low-level Color Guider}
The High-level Color Extractor can extract color information from high-level semantics. However, the color information is coarse and lacks fine-grained color guidance, making it ineffective for accurate color control. Therefore, to explore more fine-grained color features from $I_{\rm ref}$, we employ a \textbf{Low-level Color Guider (LCG)}. Specifically, our LCG consists of another trainable copy of the original DiT. We utilize the LCG to obtain intermediate representations from the reference images and concatenate them with the intermediate representations of the denoising DiT to perform self-attention:
\begin{equation}
    F_{\rm text}, F_{\rm vision} = \texttt{Self-Attention}(\texttt{concat}(F_{\rm text}, F_{\rm vision}, F_{H})),
\end{equation}
where $F_{\rm text}$ and $F_{\rm vision}$ represent the intermediate text and vision tokens of the denoising DiT.

The incorporation of LCG greatly enhances color controllability and improves the color consistency between the generated video and the reference image.

\subsection{Training Strategy}
To balance the control from sketches and the reference image, we adopt a multi-stage training strategy. We divide the training into four stages. This multi-stage training strategy not only prevents the coupling of color and geometric control but also reduces the training burden. 

In the first stage, we train the sketch-guided video generation model. Specifically, we concatenate the sketch latents with the noise as input and fine-tune the denoising DiT to enable controllable generation. This sketch conditional modeling helps accelerate convergence and improves sketch alignment. The objective function of this stage is:
\begin{equation}
\mathcal{L}_1=\mathbb{E}_{z_{0},t,c_{\rm txt},c_{\rm skt},\epsilon}[||\epsilon-\epsilon_{\theta}(z_{t},c_{\rm txt},c_{\rm skt},t)||^{2}_{2}],
\end{equation}
where $\epsilon\sim\mathcal{N}(0, 1)$ is randomly sampled from standard Gaussian noise, $t\in 1, ..., T$ denotes the diffusion timestep, $z_0$ represents the video latent encoded by VAE encoder, $z_{t}$ is the noised latent, and $c_{\rm txt}$, $c_{\rm skt}$ are the text condition and encoded sketch latents.

In the second stage, we freeze the denoising DiT and only train the Q-Former in High-level Color Extractor and the newly introduced cross attention block. This facilitates that HCE focuses on the extraction of high-level color semantic information. The objective function of this stage is:
\begin{equation}
\mathcal{L}_2=\mathbb{E}_{z_{0},t,c_{\rm txt},c_{\rm skt}, I_{\rm ref}, \epsilon}[||\epsilon-\epsilon_{\theta}(z_{t},c_{\rm txt},c_{\rm skt}, \mathcal{E}_{H}(I_{\rm ref}),t)||^{2}_{2}],
\end{equation}
where $\mathcal{E}_{H}$ is the High-level Color Extractor and $I_{\rm ref}$ indicates the reference image.

In the third stage, the denoising DiT remains frozen and we train the Low-level Color Guider. This module is designed to effectively capture fine-grained color details from the reference image for precise control. The objective function of this stage is:
\begin{equation}
\mathcal{L}_3=\mathbb{E}_{z_{0},t,c_{\rm txt},c_{\rm skt}, I_{\rm ref}, \epsilon}[||\epsilon-\epsilon_{\theta}(z_{t},c_{\rm txt},c_{\rm skt}, \mathcal{E}_{L}(I_{\rm ref}),t)||^{2}_{2}],
\end{equation}
where $\mathcal{E}_{L}$ is the Low-level Color Guider.

Finally, we freeze both the High-level Color Extractor and Low-level Color Guider. At this stage, we further finetune the denoising DiT to better integrate the color information from the two modules. This finetune greatly improves the color stability and temporal consistency of the generated animation. The objective function of this stage is:
\begin{equation}
\mathcal{L}_4=\mathbb{E}_{z_{0},t,c_{\rm txt},c_{\rm skt}, I_{\rm ref}, \epsilon}[||\epsilon-\epsilon_{\theta}(z_{t},c_{\rm txt},c_{\rm skt}, \mathcal{E}_{H}(I_{\rm ref}), \mathcal{E}_{L}(I_{\rm ref}),t)||^{2}_{2}].
\end{equation}

\begin{figure}[t]
  \centering
    \includegraphics[width=\linewidth]{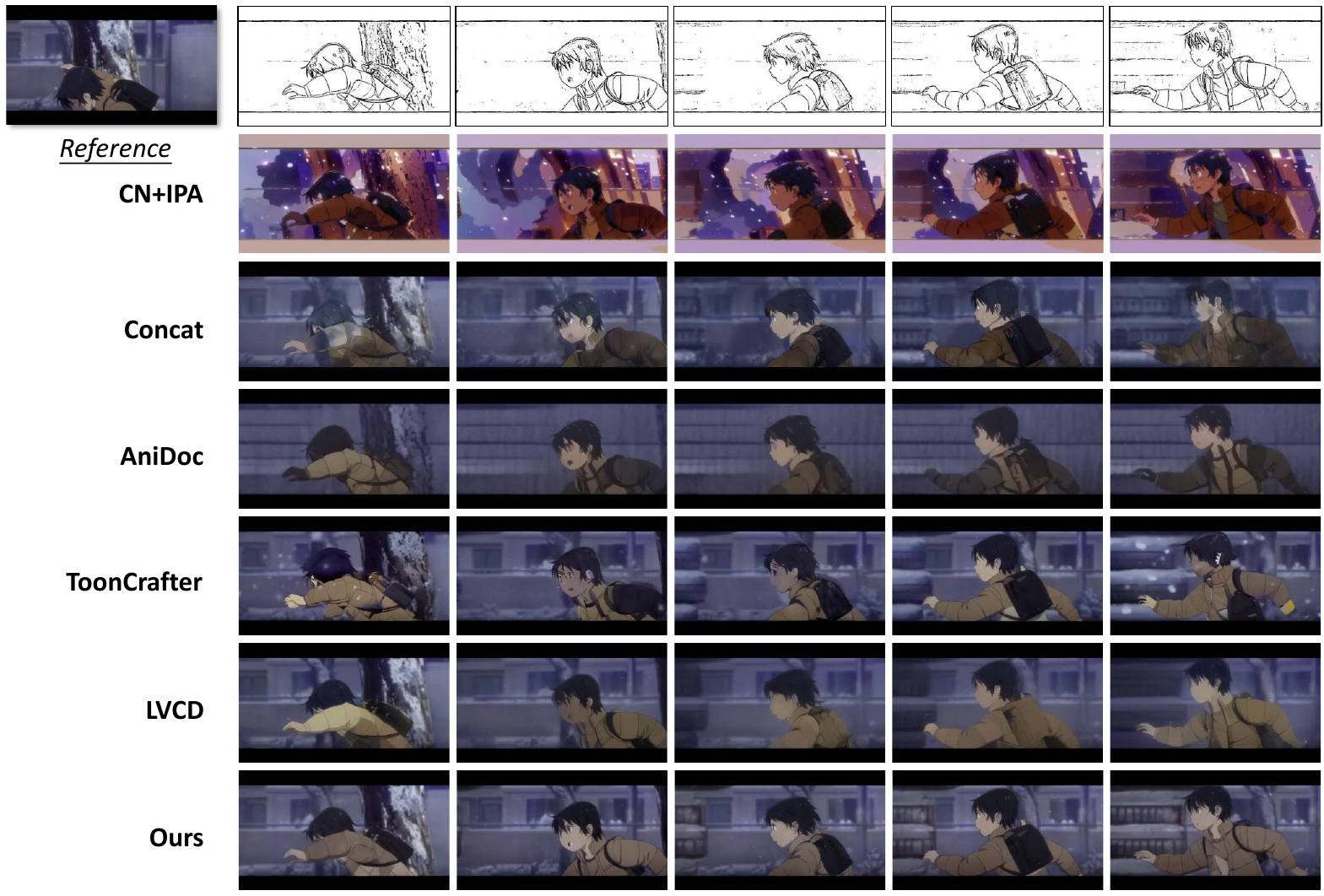}
    \includegraphics[width=\linewidth]{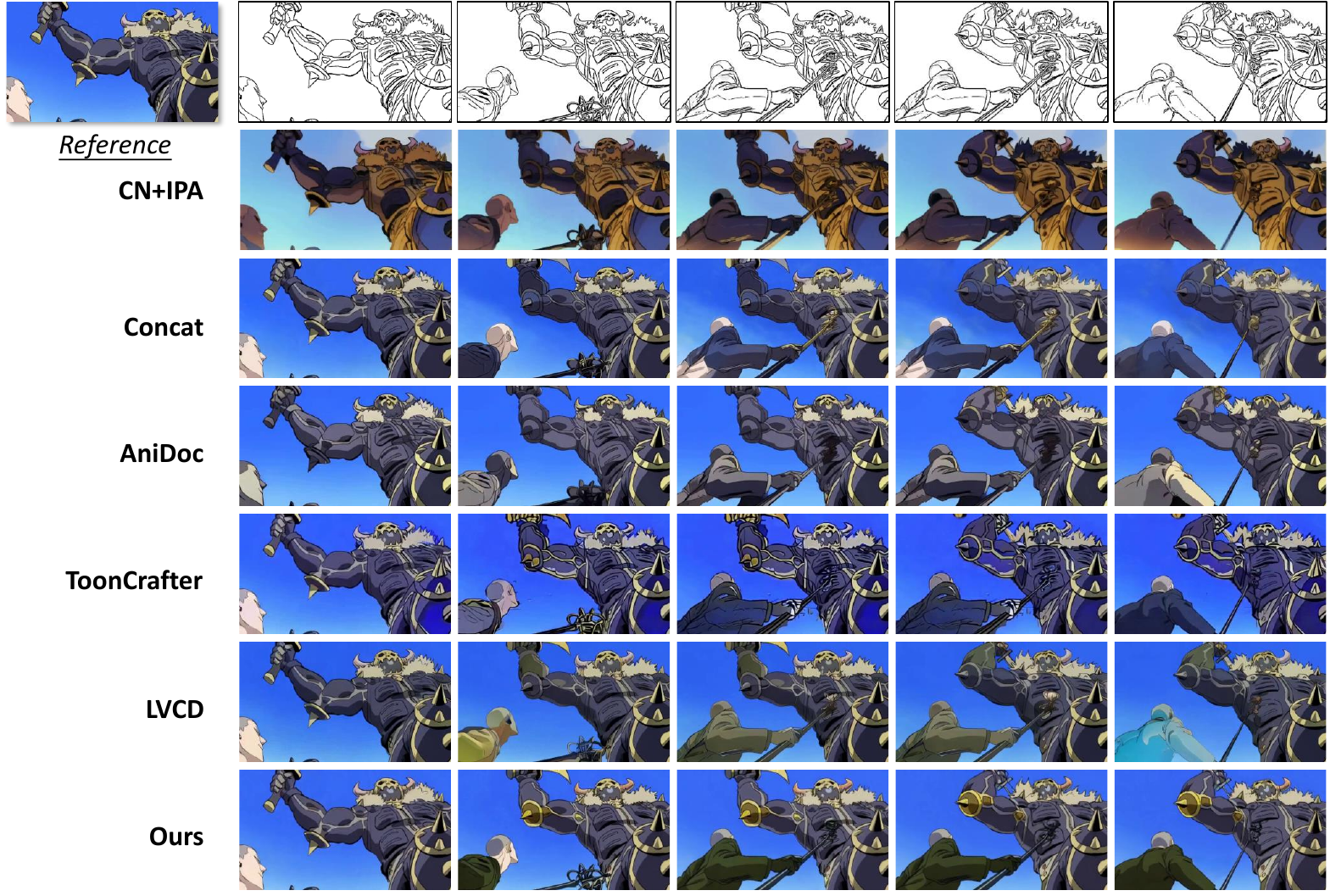}
 \vspace{-3mm}
  \caption{Qualitative comparison of reference-based colorization with CN+IPA, Concat, Anidoc, ToonCrafter, and LVCD. Our method shows great performance in terms of color accuracy, sketch alignment, temporal consistency, and visual quality.}
  \label{fig:compare}
\end{figure}

\section{Experiments}
\subsection{Experiment Setup}
\textbf{Datasets.} 
We collect a large number of anime video clips encompassing various styles and systematically curate them using OpenSora~\cite{zheng2024opensora}. We use PLLaVA~\cite{xu2024pllava} to annotate the videos as text input and utilize XDoG~\cite{winnemoller2012xdog} to obtain sketch sequences. We extract 100,000 video clips for training, with a single frame randomly selected from each clip to serve as a reference image. For the testing phase, we select another 1,800 video clips as our test set, choosing the first frame of each clip as the reference.

\noindent\textbf{Implementation Details.}
We use CogvideoX-2B~\cite{yang2024cogvideox} as our baseline and train the entire videos after grouping them into buckets based on resolution following~\cite{cogvideox-fun}. For the first stage, we train a total of 50K iterations with batch size 16. For the second stage, we train a total of 25K iterations with batch size 16. For the third stage, we train a total of 30K iterations with batch size 16. For the fourth stage, we train a total of 50k iterations with batch size 16. All experiments are performed on 8 A100 GPUs using the AdamW optimizer with a learning rate of $2 \times 10^{-5}$. We set the learnable color tokens in Q-Fomer $N$ equal to 32.

\subsection{Comparisons}
We compare our proposed AnimeColor with three existing diffusion-based reference video colorization frameworks: AniDoc~\cite{meng2024anidoc}, Tooncrafter~\cite{xing2024tooncrafter}, and LVCD~\cite{huang2024lvcd}. We also build a video colorization workflow based on AnimateDiff~\cite{guo2023animatediff} with ControlNet~\cite{zhang2023adding} and IP-Adapter~\cite{ye2023ipadapter} (denoted as CN+IPA) for comparison. What's more, based on CogvideoX-2B, we additionally train a DiT-based 
framework for comparison (denoted as Concat). This framework can also achieve reference-based animation colorization by concatenating the reference image and sketches latents with the noised latent as the input of denoising DiT.

\subsubsection{Qualitative Comparison}
We show visual examples compared with the different methods for animation colorization in Fig. \ref{fig:compare}. The two examples represent \textbf{large motion} scenes and \textbf{new objects} appear scenes. In the first example, AniDoc cannot effectively capture background features, which causes color errors. What's more, the backpack's color is not consistent with the reference. ToonCrafter's temporal domain is unstable. The character's skin color is changing. LVCD does not perform well when dealing with the large motion scene and the backpack's color changes a lot. In the second example, our method can not only keep the color of the reference part consistent with the reference image but also colorize the monk robes that do not exist in the reference image in a suitable manner. AniDoc and LVCD cannot fully capture the fine-grained color information in the reference image, such as the color of the head in the figure is not faithful to the reference image. What's more, they show poor performance in dealing with the new monk robes. ToonCrafter has obvious flickering when processing such large motions. The color of the monster on the right is also inconsistent with the reference. The results show that existing methods still have certain limitations in capturing fine-grained color control and handling large motion scenes. In addition, the Concat we trained additionally is prone to artifacts, which shows that concatenating the reference image and sketches with the noised latent for conditional generation couples the color control information in the reference with the geometric control information in the sketches, resulting in a confusing visual effect. On the contrary, our proposed method separates geometric guidance and color reference, facilitating the sketch and reference image to exert maximum guidance in controlling video generation. Overall, our approach has successfully addressed color accuracy and temporal consistency problems and achieved superior performance. 

\begin{figure}[]
  \centering
    \includegraphics[width=0.90\linewidth]{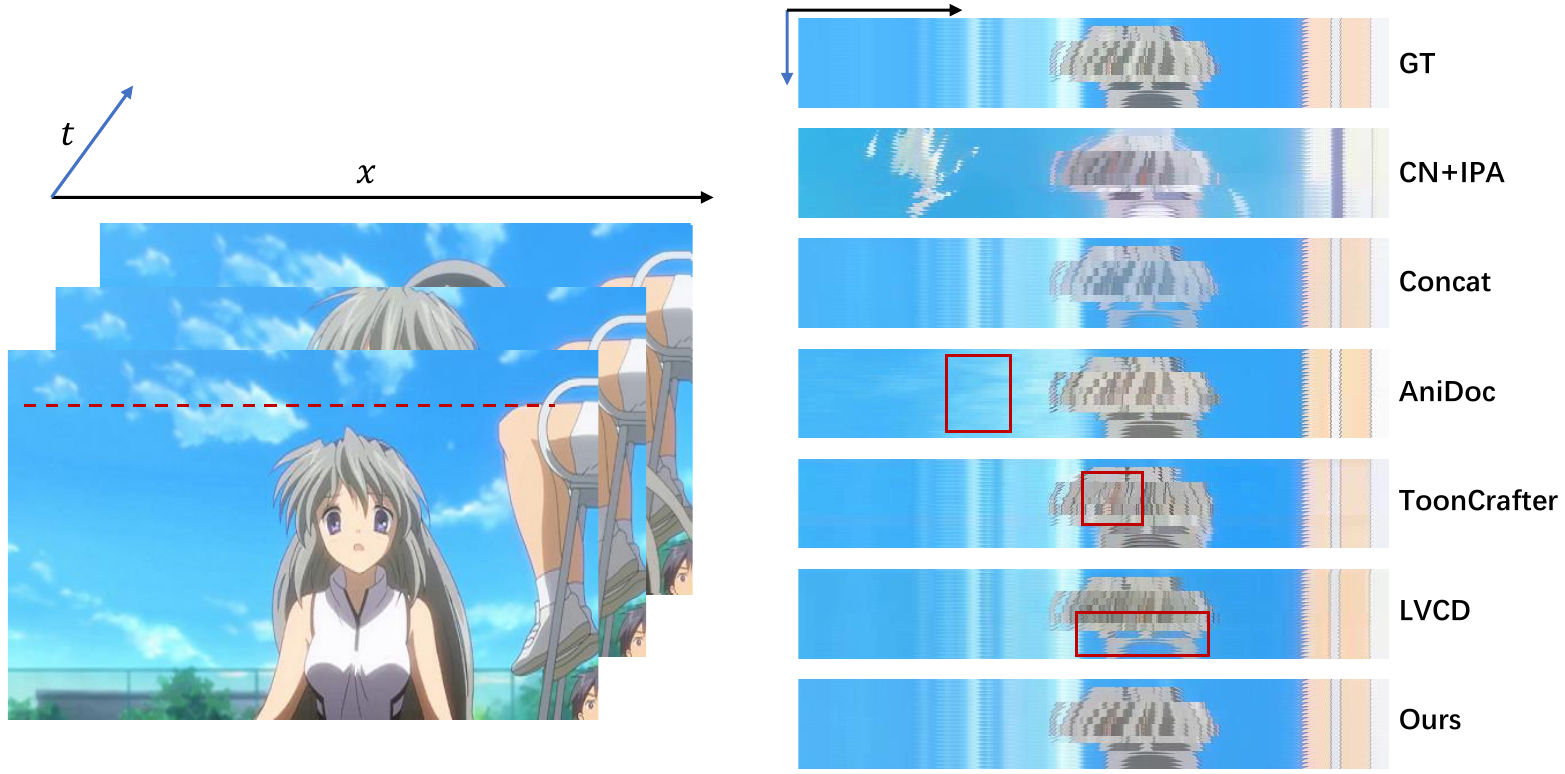}
    \vspace{-3mm}
  \caption{Temporal profile comparison of different animation colorization methods. We examine a row (shown in the red line on the left) and track changes over time (shown on the right).}
  \label{fig:temporal profile}
\end{figure}

\begin{table}
  \caption{Quantitative comparison on animation colorization with state-of-the-art methods. }
  \label{tab:compare}
  \vspace{-3mm}
  \resizebox{\linewidth}{!}{\begin{tabular}{ccccccc}
    \toprule
    Method & PSNR$\uparrow$ & SSIM$\uparrow$ & LPIPS$\downarrow$ & SA$\downarrow$ & FID$\downarrow$ & FVD$\downarrow$ \\
    \midrule
    CN+IPA & 12.33 & 0.5560 & 0.4580 & 0.3506 & 32.73 & 370.22 \\
    Concat & 20.62 & 0.7615 & 0.2584 & 0.2913 & 9.71 & 72.57 \\
    AniDoc & 16.21 & 0.6973 & 0.3351 & 0.2890 & 29.71 & 155.77 \\
    ToonCrafter & 20.44 & 0.7363 & 0.2586 & 0.3009 & 13.62 & 99.85 \\
    LVCD & 22.06 & 0.8124 & \textbf{0.2004} & 0.2758 & 9.25 & 71.63 \\
    Ours & \textbf{23.56} & \textbf{0.8288} & 0.2018 & \textbf{0.2632} & \textbf{4.87} & \textbf{23.77} \\
  \bottomrule
\end{tabular}}
\end{table}

\begin{figure}[]
  \centering
    \includegraphics[width=0.6\linewidth]{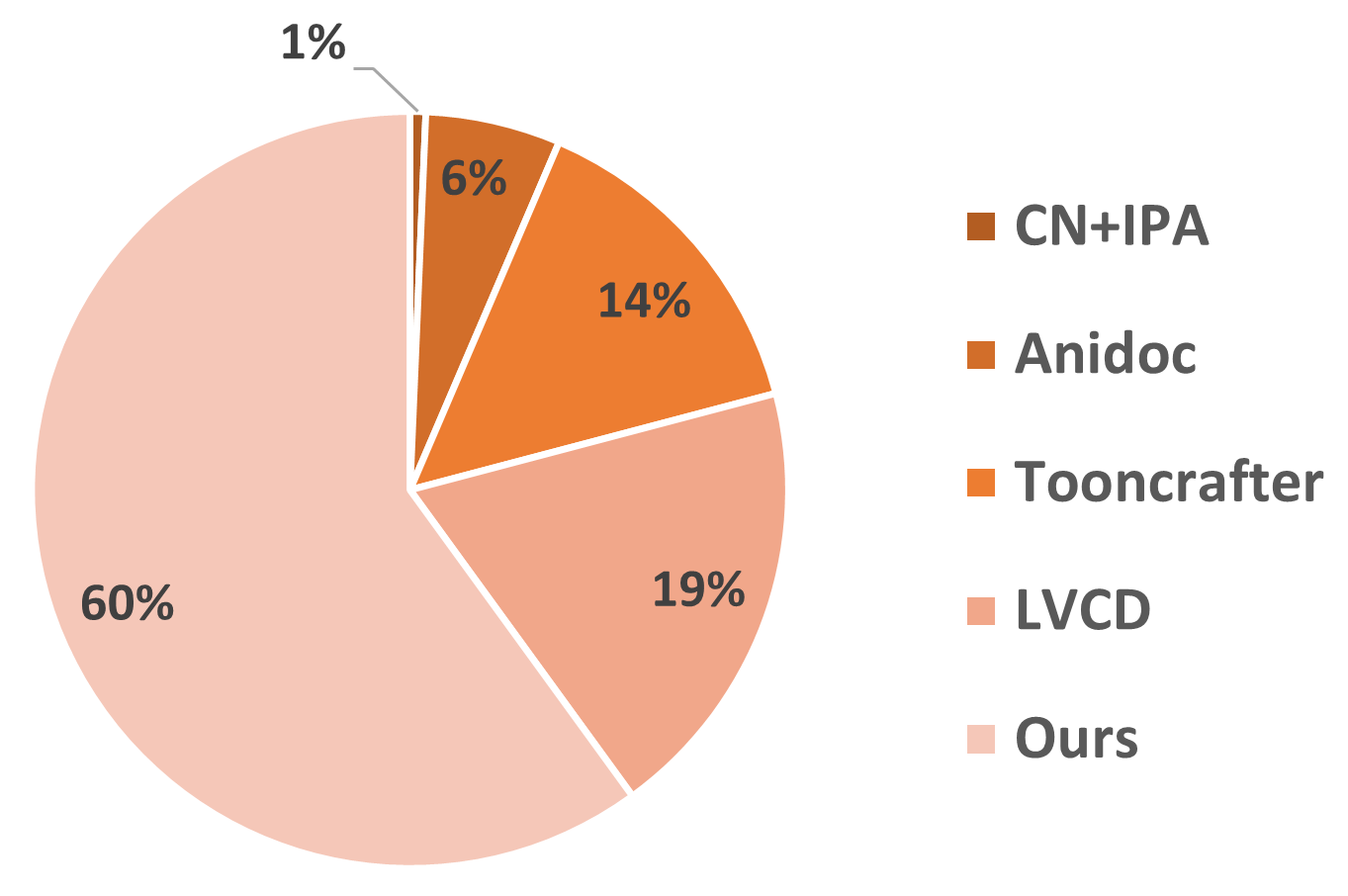}
    \vspace{-5mm}
  \caption{The results of user study. Our AnimeColor is preferred by the majority of the group.}
  \label{fig:user_study}
\end{figure}

\noindent\textbf{Temporal Consistency.} Our AnimeColor shows great performance on temporal consistency. We show a temporal profile comparison in Fig. \ref{fig:temporal profile} between our AnimeColor and other methods. CN+IPA and Concat show bad performance. AniDoc performs poorly in terms of background consistency. Tooncrafter performs poorly in terms of hair detail consistency. LVCD has color jumps in hair accessories. Our AnimeColor achieves good temporal consistency in both foreground and background.

\subsubsection{Quantitative Comparison}
We comprehensively evaluate the animation colorization methods from three aspects: (1) \textbf{Color Accuracy}: We calculate the PSNR, LPIPS, and SSIM between the generated frames and the original animation frames to evaluate the accuracy of animation colorization. (2) \textbf{Visual Quality}: We use FID~\cite{heusel2017fid} and FVD~\cite{unterthiner2019fvd} to evaluate the perceptual quality of the generated video frames and videos respectively. (3) \textbf{Sketch Alignment}: We calculate RMSE between the sketches of the generated frames and the original input sketches (SA) to evaluate whether the generated frames are aligned with the input sketches. For all metrics, we resize the frames to 512 × 320 for calculation.

We compare our methods with other state-of-the-art methods in terms of color accuracy, visual quality, and sketch alignment. As shown in Table \ref{tab:compare}, our approach outperforms other methods significantly on most metrics across three aspects. These results indicate that compared to prior works, our method is capable of generating animations with more accurate colors, superior visual quality, and better alignment with input sketches.

\subsubsection{User Study}
To further validate the effectiveness of the proposed colorization method, we conduct a user study. We invite 15 participants to watch 30 sets of videos. Each set of videos consists of a sketch sequence, colorization results from five different methods, and a reference image. Participants are asked to evaluate the colorization results comprehensively based on various aspects including video perceptual quality, color consistency with the reference image, and temporal consistency, and select the most satisfactory result. The results are shown in Fig. \ref{fig:user_study}. Our proposed AnimeColor is preferred by the majority of the group, demonstrating the significant superiority of our method compared to others and confirming the effectiveness of utilizing the controllable DiT-based video generation model in animation colorization.

\begin{table}
  \caption{Quantitative comparison about ablation study on the \textbf{model architecture}. (1): w/o HCE + LCG, (2): w/o LCG, (3): w/o HCE and (4): w/o FT. }
  \label{tab:ablation_model}
  \vspace{-3mm}
  \begin{tabular}{ccccccc}
    \toprule
    Method & PSNR$\uparrow$ & SSIM$\uparrow$ & LPIPS$\downarrow$ & SA$\downarrow$ & FID$\downarrow$ & FVD$\downarrow$ \\
    \midrule
    (1) & 14.64 & 0.7212 & 0.3442 & 0.2833 & 14.03 & 163.78 \\
    (2) & 16.35 & 0.7491 & 0.3124 & 0.2769 & 11.14 & 61.96 \\
    (3) & 20.21 & 0.8103 & 0.2291 & 0.2644 & 6.24 & 46.60 \\
    (4) & 20.89 & 0.8159 & 0.2201 & 0.2640 & 5.99 & 34.46 \\
    Ours & \textbf{23.56} & \textbf{0.8288} & \textbf{0.2018} & \textbf{0.2632} & \textbf{4.87} & \textbf{23.77} \\
  \bottomrule
\end{tabular}
\end{table}

\subsection{Ablation Studies}
\subsubsection{Ablation on Model Architecture}
\begin{figure}[]
  \centering
  \vspace{-3mm}
    \includegraphics[width=0.9\linewidth]{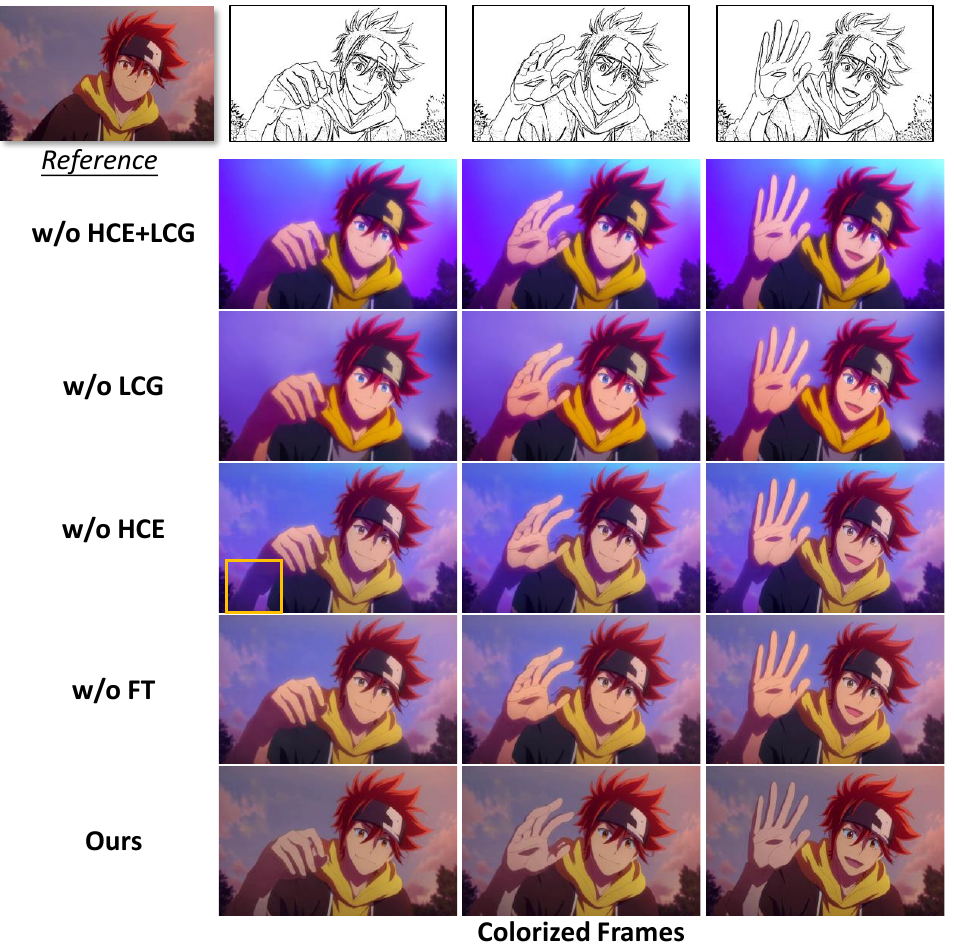}
 \vspace{-3mm}
  \caption{Qualitative comparison about ablation study on model architecture. With HCE, LCG and our training strategy, our method can achieve best performance.}
  \label{fig:ablation}
\end{figure}

To investigate the effectiveness of our proposed method, we conduct ablation studies by removing relevant modules and retraining the model. Quantitative results are presented in Table \ref{tab:ablation_model}, while qualitative results are shown in Fig. \ref{fig:ablation}. (1) w/o HCE and LCG indicates the removal of HCE and LCG, resulting in the color of the generated video being uncontrollable. What's more, FVD metric is high, which means poor visual quality. (2) w/o LCG indicates the removal of LCG, resulting in a significant decrease in color accuracy, especially PSNR. (3) w/o HCE indicates the removal of HCE. The generated video can have fine-grained color control with LCG, but lacks semantic-level color control, which can lead to problems such as color unevenness and color-semantic mismatch. Example about color error of the arm is illustrated in Fig. \ref{fig:ablation}. The (1)(2)(3) experiments verified the effectiveness of the proposed HCE and LCG modules in color control. (4) w/o FT indicates the full modules, but does not perform the fourth stage of fine-tuning. Since HCE and LCG are trained separately, directly combining the separate modules together improves the visual quality, but there is still room for improvement. After the fourth stage of fine-tuning, both qualitative and quantitative results have obvious improvement, which shows the effectiveness of the training strategy.

\begin{figure}[]
  \centering
   \vspace{-3mm}
  \includegraphics[width=0.86\linewidth]{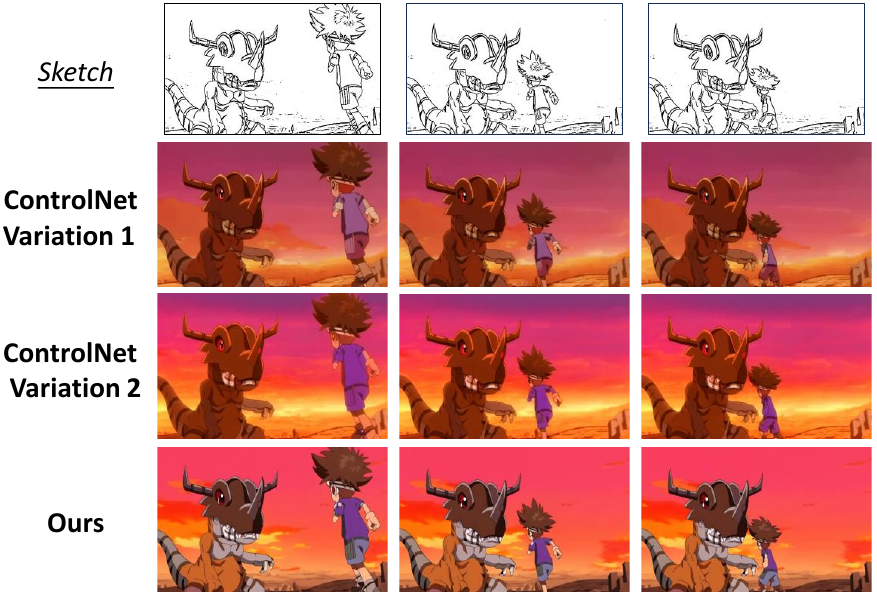}
  \vspace{-3mm}
  \caption{Qualitative comparison about ablation study on sketch conditional modeling. Our design achieves excellent performance in terms of visual quality and sketch alignment.}
  \label{fig:ablation_sketch}
\end{figure}

\begin{figure}[]
  \centering
    \includegraphics[width=0.9\linewidth]{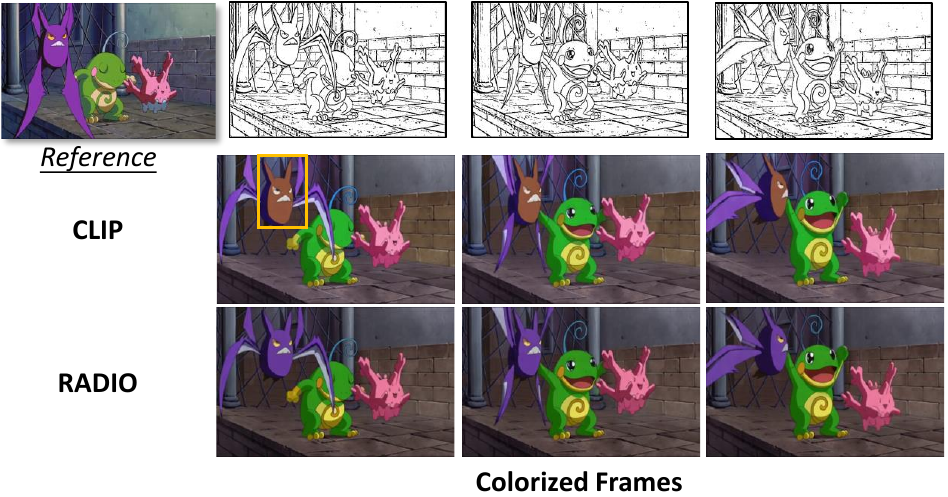}
\vspace{-3mm}
  \caption{Qualitative comparison about ablation study on image encoder in High-level Color Extractor. The RADIO-based method can solve the color unevenness problem and has more accurate color perception compared with the CLIP-based method.}
  \label{fig:ablation_encoder}
\end{figure}

\begin{table*}
  \caption{Quantitative comparison about ablation study on \textbf{sketch conditional model}. }
  \label{tab:ablation_sketch_control}
  \vspace{-3mm}
  \begin{tabular}{cccccccc}
    \toprule
    Method & PSNR$\uparrow$ & SSIM$\uparrow$ & LPIPS$\downarrow$ & SA$\downarrow$ & FID$\downarrow$ & FVD$\downarrow$ & Parameters (B) \\
    \midrule
    ControlNet Variation 1 & 13.77 & 0.6654 & 0.3834 & 0.2959 & 16.08 & \textbf{106.93} & 2.60\\
    ControlNet Variation 2 & 13.46 & 0.6507 & 0.3934 & 0.3003 & 16.20 & 136.11 & 2.60\\
    Ours & \textbf{14.64} & \textbf{0.7212} & \textbf{0.3442} & \textbf{0.2833} & \textbf{14.03} & 163.78 & \textbf{1.69}\\
    \bottomrule
  \end{tabular}
\end{table*}

\begin{table}
  \caption{Quantitative comparison about ablation study on the \textbf{image encoder} in High-level Color Extractor. }
  \label{tab:ablation_encoder}
  \vspace{-3mm}
  \begin{tabular}{ccccccc}
    \toprule
    Encoder & PSNR$\uparrow$ & SSIM$\uparrow$ & LPIPS$\downarrow$ & SA$\downarrow$ & FID$\downarrow$ & FVD$\downarrow$ \\
    \midrule
    CLIP & \textbf{16.51} & \textbf{0.7531} & 0.3145 & \textbf{0.2754} & 11.33 & 68.17 \\
    RADIO & 16.35 & 0.7491 & \textbf{0.3124} & 0.2769 & \textbf{11.14} & \textbf{61.96} \\  
  \bottomrule
\end{tabular}
\end{table}

\subsubsection{Ablation on Sketch Control}
To demonstrate the efficacy and efficiency of our proposed sketch conditional model, we evaluate several alternative conditional architectures:
(1) \textbf{ControlNet Variation 1}: Following the foundational architecture of ControlNet, we designate the first 15 Transformer blocks as the encoder in DiT and the subsequent 15 blocks as the decoder. We create trainable replicas of these encoder blocks and link each block's output to the 15 blocks of decoder with skip connections and additive operations. 
(2) \textbf{ControlNet Variation 2}: Drawing from the PixArt-$ \alpha $ ControlNet framework, we generate trainable replicas for the first 15 blocks. The output of each replica is directly connected to a zero linear layer, which is then added to the result from the corresponding frozen $i^{\text{th}}$block. The output after addition serves as the input for the next $(i+1)^{\text{th}}$ frozen block.
The quantitative and qualitative comparisons of the sketch conditional model are shown in Table \ref{tab:ablation_sketch_control} and Fig. \ref{fig:ablation_sketch}. The visualizations of qualitative comparisons demonstrate that our sketch conditional model significantly outperforms other designs in terms of visual quality. Both variations of ControlNet artificially partition the Transformer architecture into encoder and decoder components, deviating from its original design and leading to suboptimal results. Our method is better suited to the DiT architecture, enabling it to fully leverage the advantages of in-context learning, thereby significantly enhancing sketch alignment. Quantitative results presented in Table \ref{tab:ablation_sketch_control} indicate that our Sketch Conditional Modeling surpasses other methods on nearly all metrics, particularly on the SA metric. What's more, our design has fewer parameters than methods (1) and (2), facilitating easier and faster convergence during training. These results highlight the distinctive advantages of our approach in enhancing both the efficiency and the effectiveness of the modeling process.

\subsubsection{Ablation on Image Encoder in High-level Color Extractor}
In this section, we verify the superiority of RADIO over CLIP in High-level Color Extractor. We use the sketch-guided video generation model as the baseline and train the HCE with CLIP and RADIO as the image encoder respectively. The quantitative and qualitative results are shown in Table \ref{tab:ablation_encoder} and Fig. \ref{fig:ablation_encoder}, respectively. Our RADIO-based method achieves higher scores in terms of perceptual indicators such as LPIPS, FID, and FVD  while slightly lower scores in terms of PSNR, SSIM, and SA than the CLIP-based method. In addition, the CLIP-based method cannot solve the problem of color unevenness, while the RADIO-based method can solve this problem as shown in Fig. \ref{fig:ablation_encoder}. 

\begin{figure}[]
  \centering
  \vspace{-3mm}
    \includegraphics[width=\linewidth]{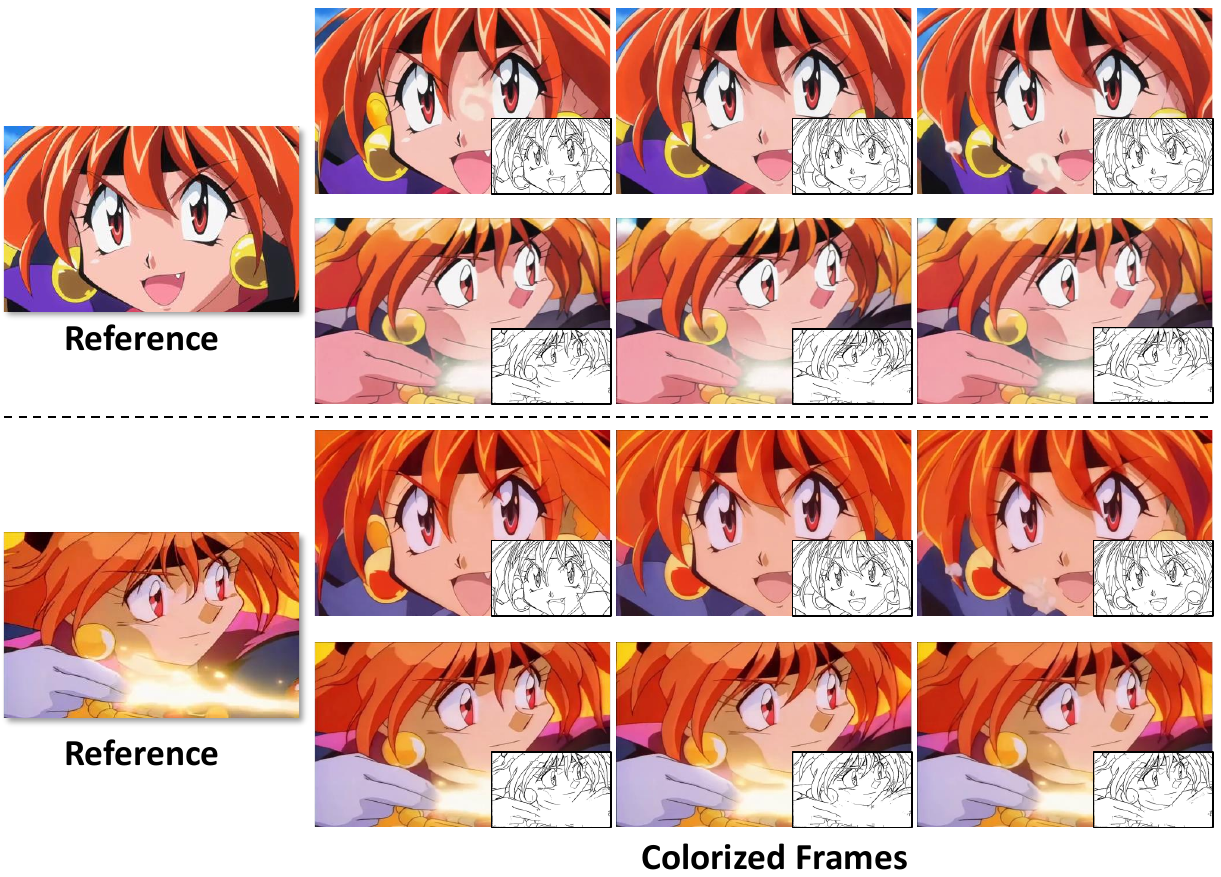}
    \vspace{-3mm}
  \caption{Illustration of the flexible application with ``Same Reference image and Different Sketches" and ``Same Sketches and Different Reference Images" settings.}
  \label{fig:application}
\end{figure}

\section{Application}
To comprehensively evaluate the versatility of our model, we extend its application to several distinct tasks, demonstrating its flexibility and robustness in diverse scenarios.

\noindent\textbf{Colorization with Same Reference and Different Sketches.}
We employ a single reference image to colorize different sketches. As illustrated in rows 1, 2 and 3,4 of Fig. \ref{fig:application}, our AnimeColor framework consistently generates colorized animation videos that maintain high color fidelity with the reference image, even when the input sketches exhibit significant variations in structure and content.

\noindent\textbf{Colorization with Same Sketches and Different Reference.}
We utilize different reference images to colorize the same sketches. As depicted in rows 1, 3 and 2, 4 of Fig. \ref{fig:application}, AnimeColor effectively adapts to the color and style of the reference images, particularly in rendering subtle details such as lighting effects and skin tones, demonstrating its ability to produce stylistically diverse outputs.

\begin{figure}[]
  \centering
    \includegraphics[width=\linewidth]{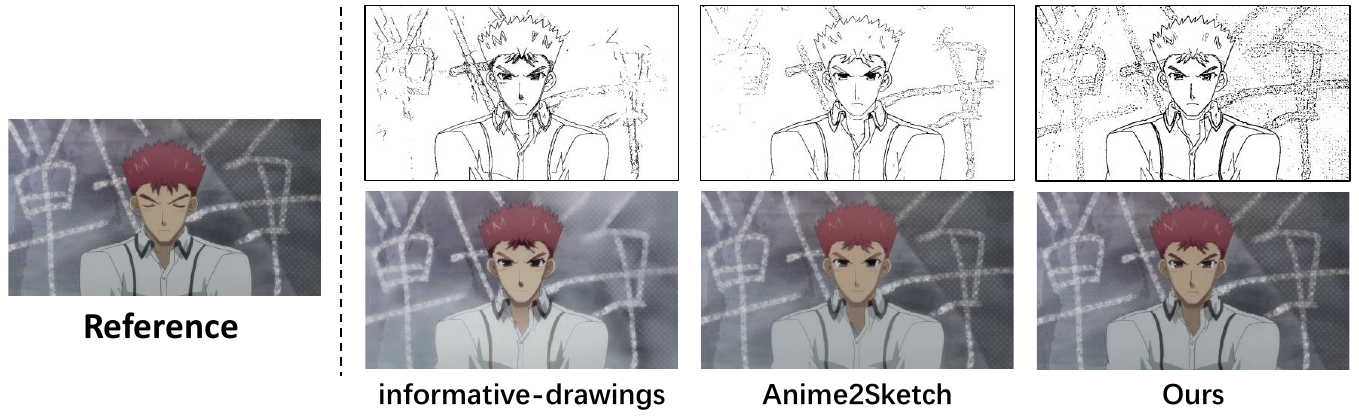}
    \vspace{-4mm}
  \caption{Impact of different lineart extraction methods.}
  \label{fig:application_sketch}
\end{figure}

\begin{figure}[]
  \centering
  \vspace{-3mm}
    \includegraphics[width=\linewidth]{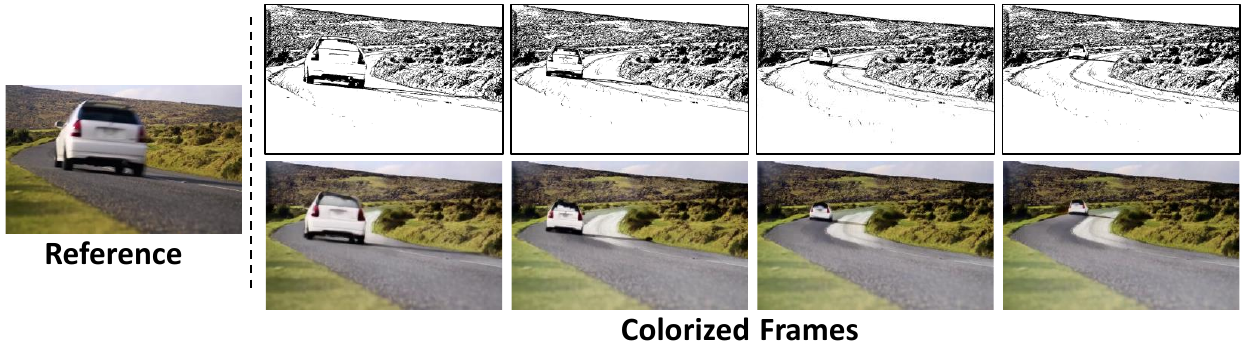}
    \vspace{-4mm}
  \caption{Colorization performance with natural images as reference.}
  \label{fig:application_real}
\end{figure}

\noindent\textbf{Colorization of Sketches from Different Line Art Extraction Methods.}
To validate the robustness of our method across varying line art extraction techniques, we evaluate its performance on sketches generated by multiple approaches. In addition to the XDoG method~\cite{winnemoller2012xdog}, we also employ informative-drawings~\cite{chan2022drawings} and Anime2Sketch~\cite{Anime2Sketch} to extract line art. As shown in Fig. \ref{fig:application_sketch}, AnimeColor successfully achieves high-quality colorization while maintaining consistency with the reference image, despite minor variations in detail caused by differences of the extraction methods.

\noindent\textbf{Colorization with Natural Images as Reference.}
To further assess the generalization capability of our approach, we conduct experiments using natural images as reference inputs. As demonstrated in Fig. \ref{fig:application_real}, our method exhibits robust colorization performance even when the reference images are natural scenes. The results highlight the model's ability to produce visually consistent and high-quality colorizations across a wide range of reference image types, underscoring its adaptability and practical utility.

\section{Conclusion}
In this paper, we propose AnimeColor, a DiT-based framework for reference-based animation colorization. AnimeColor addresses the challenges faced by existing colorization methods, particularly in handling scenes with large motion, such as camera movements and dynamic character actions. We integrate sketch control with the in-context learning capability of DiT. To achieve accurate colorization, we design two complementary modules: a High-level Color Extractor for semantic color consistency and a Low-level Color Guider for fine-grained color control, both of which extract and integrate color guidance from reference images. Furthermore, we propose a multi-stage training strategy to ensure the effective optimization of each module. Extensive experiments demonstrate the superiority of AnimeColor over existing methods in terms of color accuracy, sketch alignment, temporal consistency, and visual quality.

\begin{acks}
This work was partly supported by the NSFC62431015, the Fundamental Research Funds for the Central Universities, Shanghai Key Laboratory of Digital Media Processing and Transmission under Grant 22DZ2229005, 111 project BP0719010.
\end{acks}

\bibliographystyle{ACM-Reference-Format}
\bibliography{reference}

\end{document}